\documentclass{jdmdh}
\usepackage{array}
\usepackage{pgfplots}
\usepackage[T1]{fontenc}

\usepackage[utf8]{inputenc}
\usepackage{arabtex}
\usepackage{utf8}
\setcode{utf8}

\pgfplotsset{compat=1.18}

\titlespacing*{\section}
{0pt}{2.5ex plus 1ex minus .2ex}{1.3ex plus .2ex}
\titlespacing*{\subsection}
{0pt}{1.0ex plus 1ex minus .2ex}{1.3ex plus .2ex}
\titlespacing*{\subsubsection}
{0pt}{1.0ex plus 1ex minus .2ex}{1.3ex plus .2ex}

\title{Predicting Sustainable Development Goals Using Course Descriptions - from LLMs to Conventional Foundation Models}
\author[1]{Lev Kharlashkin, Melany Macias, Leo Huovinen and Mika Hämäläinen}
\affil[1]{Metropolia University of Applied Sciences, Finland}

\corrauthor{Lev Kharlashkin}{lev.kharlashkin@metropolia.fi}

\begin{document}

\maketitle

\abstract{We present our work on predicting United Nations sustainable development goals (SDG) for university courses. We use an LLM named PaLM 2 to generate training data given a noisy human-authored course description input as input. We use this data to train several different smaller language models to predict SDGs for university courses. This work contributes to better university level adaptation of SDGs. The best performing model in our experiments was BART with an F1-score of 0.786.}


\section{Introduction}


The United Nations (UN) has established a list of 17 sustainable development goals (SDGs)\footnote{https://sdgs.un.org/goals}. These goals are becoming more and more important in understanding the societal, humanitarian and environmental impact of companies in the EU given that certain large companies need to take rigorous sustainability reporting as part of their annual reporting to the authorities\footnote{https://finance.ec.europa.eu/capital-markets-union-and-financial-markets/company-reporting-and-auditing/company-reporting/corporate-sustainability-reporting\_en}.

Because of the growing importance, many universities and educational institutions have started to adopt the UN SDGs as part of their academic curricula. This raises the important question of how an educational institution can, on a higher level, know which SDGs are being taught and where in the different degree programs.

Adopting local models that are adjusted based on their course descriptions helps universities to comply with GDPR\footnote{https://eur-lex.europa.eu/eli/reg/2016/679/oj} and preserve data privacy. Additionally, by tailoring these models to the unique linguistic and curriculum quirks of the school, prediction accuracy can be increased while maintaining the security and confidentiality of sensitive data.

In our paper, we collect and clean a noisy course description dataset and use an LLM to generate SDGs for each course. We manually check and fix the LLM generated data that is used for testing. Furthermore, we fine-tune several smaller foundation models to predict SDGs based on course descriptions. Fine-tuning a smaller model makes the SDG prediction task faster and more cost-efficient.

\section{Related work}

Sustainable development has been studied in the field of NLP from many different points of view such as studying fairness in NLP \citep{hessenthaler-etal-2022-bridging}, studying poverty and societal sustainability in interviews \citep{van-boven-etal-2022-intersection}, argumentation mining \citep{fergadis-etal-2021-argumentation} and community profiling \citep{conforti-etal-2020-natural} among others. Our take differs from these in the sense that we aim to cover all UN sustainable development goals and apply them in a pedagogical context.

Perhaps the most similar prior work to ours is that of \citet{amel2021nlp}. They used more traditional methods such as word2vec \citep{mikolov2013efficient} and doc2vec \citep{le2014distributed} to assess how well companies align with UN SDGs. They use a dictionary of SDG goal related terms to assess the overlap of each SDG with a given company. They then train a logistic classifier, an SVM and a fully connected neural network on the embeddings. Their finding was that a combination of doc2vec and SVM gave the best results.

In terms of the pedagogical context of our research, there is plenty of prior research on incorporating SDGs as part of teaching \citep{collazo2020implementation,rajabifard2021applying,kwee2021want}. This prior research is non-computational and to best of our knowledge, there is no prior research work on the topic from the NLP stand point. 

\section{Data}

For our work, we gathered course information from Metropolia University of Applied Sciences over their API\footnote{https://wiki.metropolia.fi/display/opendata/REST-rajapinnat}. This data retrieved was composed of 51,386 Finnish and English courses from the years 2004 to 2023.

\begin{figure}[ht]
\centering
\includegraphics[width=11cm]{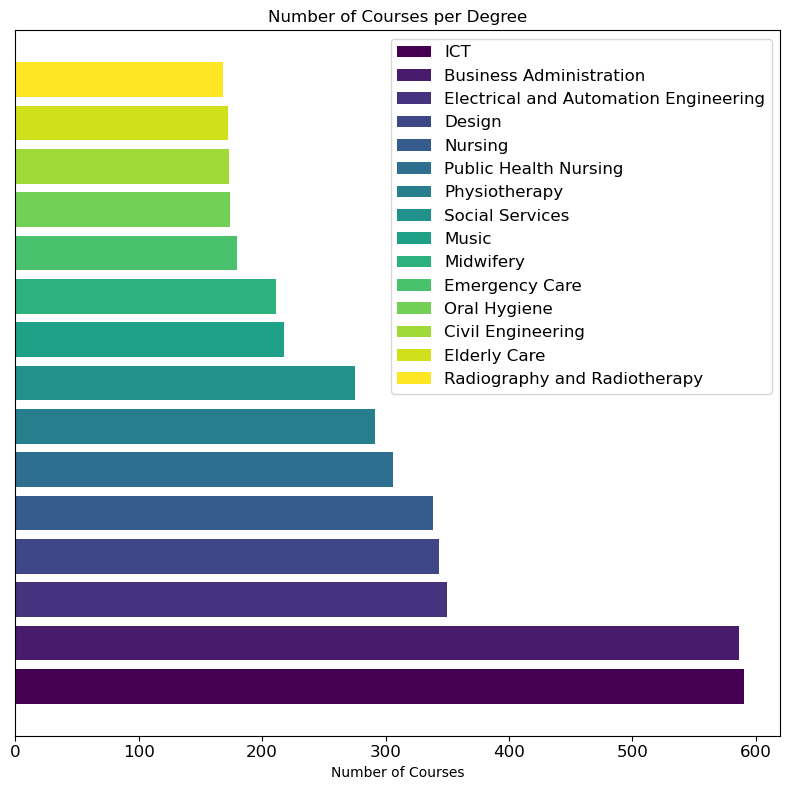}
\caption{Distribution of courses per degree after the initial cleaning step.}
\label{fig:courses_per_degree}
\end{figure}

\subsection{Data Preprocessing}

The dataset presented significant challenges in terms of variability and noise, attributed to the subjective nature of course descriptions provided by individual instructors. The length of these descriptions varied widely, and in some cases, the course objectives were either missing or contained redundant or extraneous information.

The study focused on courses offered between 2021 and 2023 to capture recent curricular trends. We imposed a character limit of 500 to 2000 for the combined length of course descriptions and objectives to maintain an optimal balance of detail and brevity. Courses outside this range were excluded. Moreover, the study was confined to courses conducted in English to maintain consistency in language processing. At this point, the data was composed of 8708 courses, with 103 unique disciplines.

Figure \ref{fig:courses_per_degree} depicts the distribution of English courses per the top 15 degrees after the initial cleaning step, which illustrates the diverse curricular offerings within the analyzed period. Notably, the 'Information and Communication Technology' discipline demonstrates a significantly higher volume of courses, underscoring the sector's expansion and its pivotal role in contemporary education landscapes. This visual representation also serves to highlight the curricular focus areas that are apparent within the institution, guiding the subsequent analysis stages to probe into the qualitative aspects of course content more deeply.

Our standardization process involved several steps, specifically the removal of entries with missing course descriptions or objectives, language detection using the Spacy NLP library \citep{spacy2} to remove Finnish courses and retain only English courses, and elimination of duplicates present from courses offered in multiple years.

The dataset, in its finalized state, consisted of 2125 courses in English, each defined by three key elements: name, description, and objective.

\subsection{Generating SDGs}

We use PaLM 2 \citep{anil2023palm} over Vertex AI API \footnote{https://cloud.google.com/vertex-ai/docs/reference/rest} to generate the training data for the SDG prediction models. In particular, we use \textit{text-bison-32k} from Model Garden\footnote{https://cloud.google.com/vertex-ai/docs/generative-ai/learn/models}. PaLM 2 is an LLM that takes in a prompt and produces an output based on the prompt in a similar fashion to ChatGPT \citep{openai2023gpt4}.

In search of the most effective prompt, we employed the prompting IDE tool Prompterator \citep{sucik-etal-2023-prompterator}. Thus, to ensure the quality of the model's outputs, we took a small sample of data for batch processing and manually reviewed the model's responses using Prompterator. 

This evaluation helped us confirm the appropriateness of the SDG predictions for subsequent training. Moreover, batch processing was instrumental in handling the dataset efficiently, allowing for the dynamic integration of each course's metadata into the prompt template. The responses collected from the model included the SDG goals deemed most relevant by the LLM, as shown in Table \ref{tab:final_prompt_spec}.

\begin{table}[]
\centering
\small
\begin{tabular}{|l|p{5.5cm}|}
\hline
\textbf{Parameter} & \textbf{Value} \\
\hline
Prompt & Your goal is to identify UN SDG Goals relevant to students. Given a {course name}, the student learns: {course content} and {course objective}. Answer the question: What are the top few most relevant sustainable development goals to this course? Your task is to return only the numbers of the top few goals separated by commas. Also, never use the goal number 4. \\
\hline
Temp. & 0.2 \\
\hline
Token Limit & 500 \\
\hline
\end{tabular}
\caption{Final prompt specification used for SDG goal generation}
\label{tab:final_prompt_spec}
\end{table}

Our final prompt to the model was the following one appended with a course description of each course: \textit{Your goal is to identify UN SDG Goals relevant to students. Given a {course name}, the student learns: {course content} and {course objective}. Answer the question: What are the top few most relevant sustainable development goals to this course? Your task is to return only the numbers of the top few goals separated by commas. Also, never use the goal number 4.}

The purpose of selecting a lower temperature setting (0.2) was to limit the output variability of the model and promote accuracy. We discovered empirically that a token limit of 500 was adequate to enable the model to produce thorough answers without being overly verbose.

\subsection{Data Preparation for Training}

After SDG predictions were generated by the  LLM, the dataset underwent a meticulous cleaning process. Initially, labels generated by the language model were extracted, stripping away the prompts included in the input. Subsequently, we refined the labels to solely represent SDG numbers, with a particular exclusion of Goal 4: Quality education. This goal was excluded because it was over-represented in the data - virtually every single university course contributes to quality education.

For compatibility with multi-label classification models, we encoded the list of SDGs relevant to each course into a binary format. Consequently, the dataset for model training comprised two components: the input—encompassing the course name, description, and objectives—and the output—a binary vector denoting pertinent SDGs.

\begin{figure}[ht]
\centering
\includegraphics[width=\textwidth]{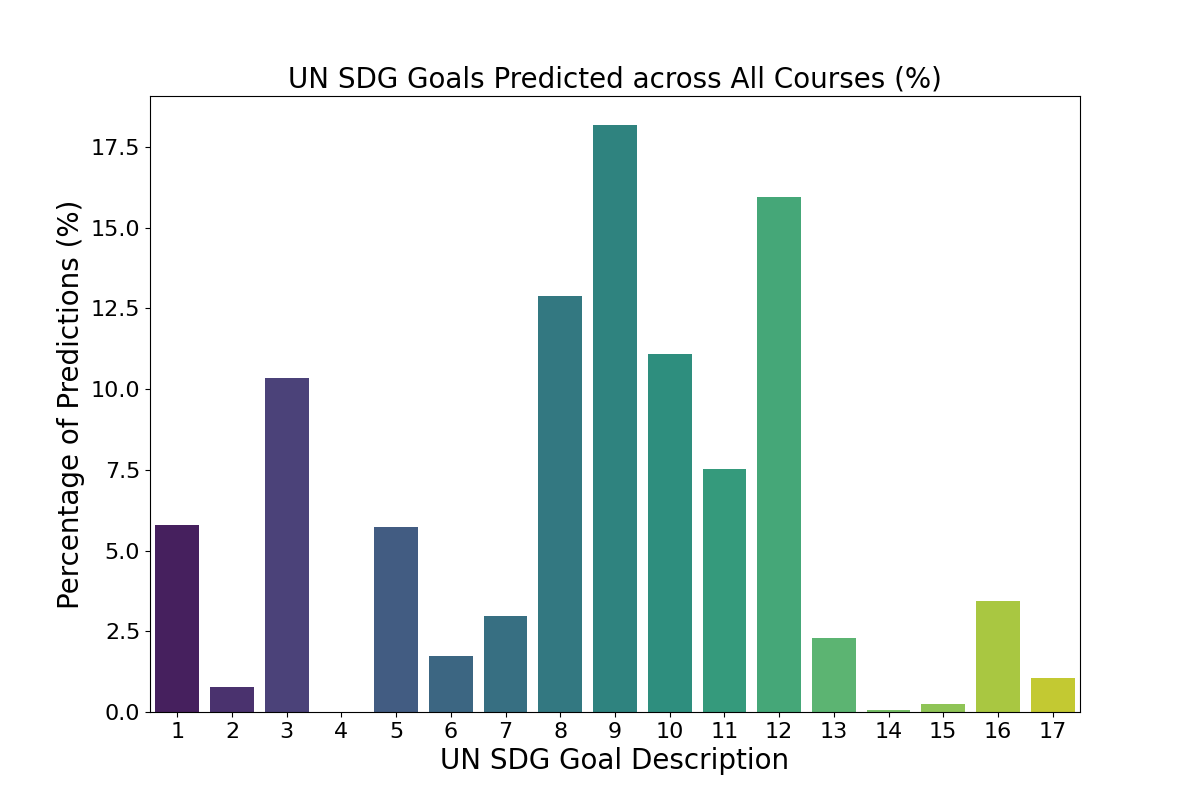}
\caption{Distribution of SDG mentions within the training dataset.}
\label{fig:sdg_distribution}
\end{figure}

An evaluation of the SDG distribution within the training data was conducted to ascertain dataset quality and representation balance, the results of which are depicted in Figure \ref{fig:sdg_distribution}.

The percentage distribution of the model-generated SDG forecasts is shown in Figure \ref{fig:sdg_distribution}.  This figure omits Goal 4 (Quality Education) by design and reveals that certain goals, such as 2 (Zero Hunger), 14 (Life Below Water), and 15 (Life on Land), are less frequently associated with the course descriptions at Metropolia. This skewness in the data reflects the varied emphasis of SDGs in the actual course content.

After the quality of the dataset was confirmed, it was split 70:15:15 into subsets for training, validation, and testing. This allocation prevents overfitting during training and enables a thorough assessment of the model's performance across unknown data.

\section{SDG prediction models}

We try out fine-tuning several models for multi-class classification task using Transformers Python library \citep{wolf-etal-2020-transformers}. We selected BERT \citep{devlin-etal-2019-bert}, mBERT \citep{devlin-etal-2019-bert}, RoBERTa \citep{liu2019roberta}, XLM-RoBERTa \citep{conneau2020unsupervised}, and BART \citep{lewis2019bart} due to their top-tier results in multi-label classification tasks. XLM-RoBERTa and mBERT, in particular, were chosen to explore theirs capabilities of a multilingual model for potential future purposes.

The models are trained to receive course information as input and are trained to predict the corresponding 3 most relevant SDG goals as output. This aligns with the format of our training data. An illustrative example of the input data and the model's expected output is presented in Table \ref{tab:input_output_example}. This binary output represents the relevance of specific SDG goals to the given course information, with the example indicating relevance to goals 3, 5 and 8.

\begin{table}[ht]
\centering
\footnotesize
\begin{tabular}{|p{0.2\columnwidth}|p{0.8\textwidth}|}
\hline
\textbf{Field} & \textbf{Value} \\
\hline
Input & "Clinical Practice, the student learns: Clinical Practice in nursing environment. Students- can apply the theoretical and clinical competence required by the clinical practice environment to the nursing care of clients/patients- can maintain and promote the health of clients/patients and their significant others in a client-oriented way in nursing care- follow the ethical guidelines and principles of nursing- work responsibly as members of work groups and work community- can assess their professional competence and develop it further." \\
\hline
Output & [0, 0, 1, 0, 1, 0, 0, 1, 0, 0, 0, 0, 0, 0, 0, 0, 0] \\
\hline
\end{tabular}
\caption{Example of course information input and the expected binary output for SDG prediction.}
\label{tab:input_output_example}
\end{table}

Training and evaluation of the models were carried out on Puhti, a Finnish research supercomputer provided by CSC - IT Center for Science, which facilitated the necessary computational resources \citep{csc2023puhti}. Using the V100 GPU's large memory and parallel processing power, the models were trained on a single node to effectively handle our dataset.

\begin{figure}
\centering
\includegraphics[width=\textwidth]{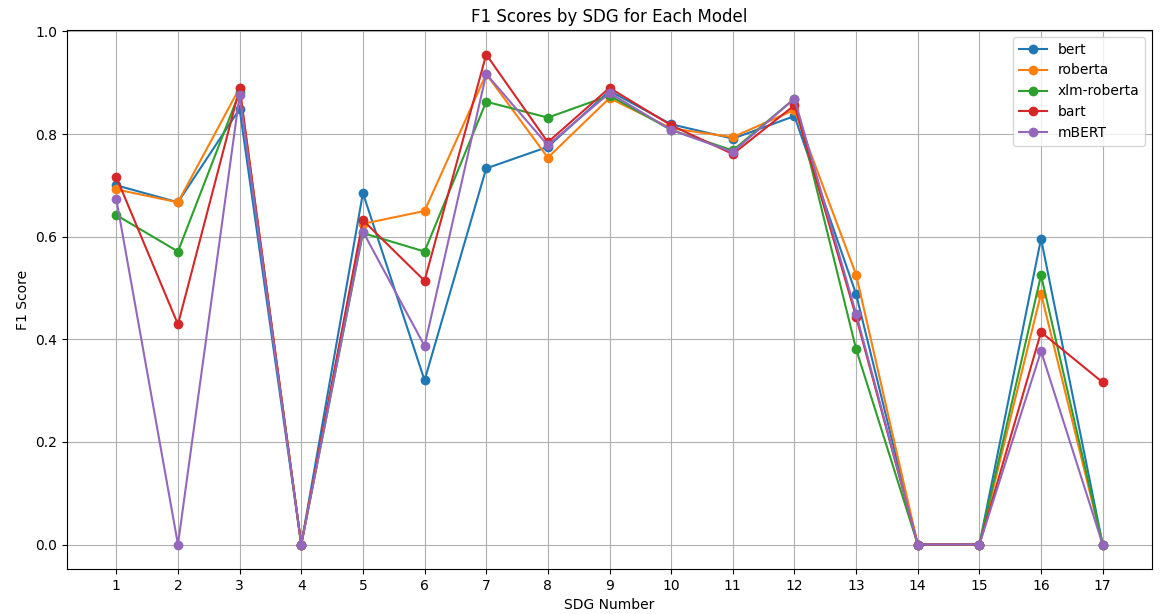}
\captionof{figure}{F1 Scores by SDG for Each Model}
\label{fig:f1_sdg_per_model}
\end{figure}

\section{Results}

Understanding the efficacy of a given algorithm in the context of multi-class SDG classification depends on the evaluation of model performance. The models' performance was evaluated using precision, recall, and F1-score, which offer a thorough understanding of the models' capabilities—especially in light of the inherent class imbalance in our dataset. The performance metrics for each model are shown in the following table, emphasizing their advantages and disadvantages for the given task.

\begin{table}[ht]
\centering
\begin{tabular}{@{}lccc@{}}
\hline
Model & Precision & Recall & F1-Score \\
\textbf{BERT}        & 0.765 & 0.798 & 0.781 \\
\textbf{mBERT}       & 0.762 & 0.795 & 0.778 \\
\textbf{RoBERTa}     & 0.768 & 0.802 & 0.785 \\
\textbf{XLM-RoBERTa} & 0.767 & 0.801 & 0.784 \\
\textbf{BART}        & \textbf{0.769} & \textbf{0.803} & \textbf{0.786} \\
\hline
\end{tabular}
\caption{Models performance based on the micro scores}
\label{tab:models_micro_scores}
\end{table}

The table  \ref{tab:models_micro_scores} shows BERT's precision of 0.765 and F1-score of 0.781 reflect its proficiency in categorizing instances correctly, demonstrating a reliable balance between precision and recall.

In contrast, BART outperforms other models with the highest F1-score, suggesting superior model efficacy due to its advanced pretraining methodology. The nuanced performance variations across the models underscore the significance of model selection tailored to specific NLP tasks' requirements.

The displayed F1 scores reveal that model performance fluctuates across the SDGs, with BART and mBERT often outperforming others, particularly in SDGs 7, 8, and 9. The lower F1 scores for SDGs 14,15 and 17 suggest that data imbalances pose challenges, affecting the models' ability to generalize effectively in these areas. Such patterns indicate the need for enhanced data strategies to address the imbalances and optimize model performance.


\section{Conclusions}


In this paper, we introduced a novel approach to predicting UN SDGs for university courses, employing PaLM 2 large language model to generate training data from course descriptions. Through the utilization of various smaller language models, we successfully trained models to predict SDGs for university courses. Notably, the best-performing model in our experiments was BART, achieving an F1-score of 0.786. 

This research contributes to advancing the integration of SDGs at the university level, providing a valuable methodology for enhancing the adaptation of sustainable development principles in higher education. The findings open avenues for further research and implementation of similar approaches to foster sustainable practices in academic institutions worldwide.

\bibliographystyle{plainnat}
\bibliography{custom}

\appendix\footnotesize

\end{document}